\documentclass[10pt,journal,twoside]{IEEEtran}

\usepackage{graphicx}
\usepackage{color}
\usepackage{placeins}
\usepackage{float}
\usepackage{tabularx,colortbl}
\usepackage{multirow}
\usepackage{makecell}
\usepackage{CJK}
\usepackage{amsmath,bm}
\usepackage{cite}
\usepackage{amsthm}
\usepackage{extarrows}
\usepackage{amssymb}
\usepackage[table]{xcolor}
\usepackage{amsmath}

\begin{document}

%%%%%%%%% TITLE

\title{Just Noticeable Difference for Deep Machine Vision}

\author{
	Jian Jin,~\IEEEmembership{Member,~IEEE},
	Xingxing Zhang,
	Xin Fu,
	Huan Zhang,
	
	Weisi Lin,~\IEEEmembership{Fellow,~IEEE},
	Jian Lou, 
	Yao Zhao,~\IEEEmembership{Senior Member,~IEEE}
	
	\thanks{This work was supported by Alibaba Group through Alibaba Innovative Research (AIR) Program and Alibaba-NTU Singapore Joint Research Institute (JRI), Nanyang Technological University, Singapore. \emph{(Corresponding author: Weisi Lin.)}}
	\thanks{J. Jin, H. Zhang, and W. Lin are with the School of Computer Science and Engineering, Nanyang Technological University, 639798, Singapore. J. Jin and W. Lin are also with Alibaba-NTU Singapore Joint Research Institute, Nanyang Technological University, 639798, Singapore. E-mail: jian.jin@ntu.edu.sg, huan.zhang@siat.ac.cn, wslin@ntu.edu.sg.}
	\thanks{X. Zhang is with the Department of Computer Science and Technology, Tsinghua University, Beijing 100084, China. E-mail: xxzhang2020@mail.tsinghua.edu.cn.}
	\thanks{X. Fu and Y. Zhao are with the Institute of Information Science, Beijing Jiao Tong University, Beijing 100044, China, and also with the Beijing Key Laboratory of Advanced Information Science and Network Technology, Beijing 100044, China. E-mail: \{xinfu and yzhao\}@bjtu.edu.cn.}
	\thanks{J. Lou is with the Alibaba cloud business group, department of video cloud, Alibaba, Hangzhou 310052, China. Email: jianedwardlou@gmail.com.}
%	\thanks{Copyright \copyright 20XX IEEE. Personal use of this material is permitted. However, permission to use this material for any other purposes must be obtained from the IEEE by sending an email to pubs-permissions@ieee.org.}
%	\thanks{This work was} 
}

% The paper headers
\markboth{IEEE Transactions on Circuits and Systems for Video Technology.}{Jin {\it \lowercase{et al.}}: {Just Noticeable Difference for Deep Machine Vision}
}

\maketitle

%%%%%%%%% ABSTRACT
\begin{abstract}
As an important perceptual characteristic of the Human Visual System (HVS), the Just Noticeable Difference (JND) has been studied for decades with image and video processing (e.g., perceptual visual signal compression). However, there is little exploration on the existence of JND for the Deep Machine Vision (DMV), although the DMV has made great strides in many machine vision tasks. In this paper, we take an initial attempt, and demonstrate that the DMV has the JND, termed as the DMV-JND. We then propose a JND model for the image classification task in the DMV. It has been discovered that the DMV can tolerate distorted images with average PSNR of only 9.56dB (the lower the better), by generating JND via unsupervised learning with the proposed DMV-JND-NET. In particular, a semantic-guided redundancy assessment strategy is designed to restrain the magnitude and spatial distribution of the DMV-JND. Experimental results on image classification demonstrate that we successfully find the JND for deep machine vision. Our DMV-JND facilitates a possible direction for DMV-oriented image and video compression, watermarking, quality assessment, deep neural network security, and so on. 
\end{abstract}

\begin{IEEEkeywords}
	Just noticeable difference (JND), human visual system (HVS), deep  machine vision (DMV), image classification, class activation mapping (CAM)  
\end{IEEEkeywords}

%%%%%%%%% BODY TEXT
\section{Introduction}
\label{Intro}
% \IEEEPARstart{T}{he} unique psychological and physiological mechanisms of the Human Visual System (HVS) make humans unable to perceive certain changes in images and videos due to their underlying spatial-temporal sensitivities and masking properties \cite{jayant1993signal}. 
\IEEEPARstart{T}{he} unique psychological and physiological mechanisms of the Human Visual System (HVS) make humans unable to perceive certain changes in images and videos. This is due to its underlying spatial-temporal sensitivities and masking properties \cite{jayant1993signal}. 
% That is, images and videos have visual redundancy for the HVS. The HVS oriented Just Noticeable Difference (JND), termed as HVS-JND, refers to find the maximum visual threshold of each pixel. Any changes under the threshold can be tolerated by the HVS (i.e., the homogeneous property of JND). The homogeneous property makes HVS-JND widely used in image and video processing, 
That is, images and videos have visual redundancy for the HVS. The HVS oriented Just Noticeable Difference (JND), termed as the HVS-JND, refers to find the maximum visual threshold of each pixel. Any changes under the threshold can be tolerated by the HVS. Commonly, this kind of property of JND is regarded as the homogeneous property, which exists in human perception, such as vision, hearing, smell, touch, taste, and so on. All changes below JND form a homogeneous range that leads to the same perception.  
\begin{figure}[htbp]
	\begin{center}
		\noindent
		\includegraphics[width = 3.35 in]{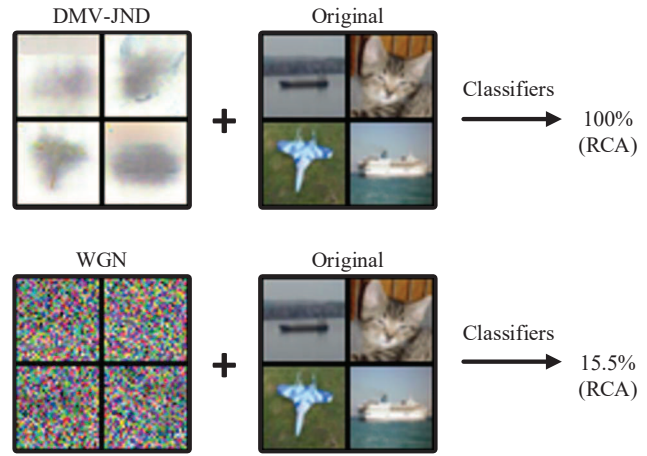}
		\caption{The Relative Classification Accuracy (RCA) comparison between DMV-JND distorted image and White Gaussian Noise (WGN) distorted image. After adding DMV-JND (generated via our proposed DMV-JND model) and WGN (with same amount of noise) to the original image, we get 100\% and 15.55\% RCA on the CIFAR-10 dataset, respectively.}\label{RCA}
		%\caption{The RCA comparison between DMV-JND distorted image and random-noise-distorted image. After adding DMV-JND (generated via our proposed DMV-JND model) and WGN to the original image, we get 100\% and 15.55\% RCA on the CIFAR-10 dataset, respectively.}\label{RCA} 
	\end{center}
\end{figure}
The homogeneous property reflects the characteristics in sensitivity of the human perception, which makes the HVS-JND being widely used in image and video processing, such as perceptual visual signal compression \cite{jayant1993signal}, quality-of-experience (QoE) in video streaming service \cite{zhang2018deepqoe}, watermarking \cite{chou2010perceptually}, error resilience \cite{karam2007selective}, supper resolution \cite{karam2011efficient}, graphic rendering \cite{nader2015just}, and so on.

%RCA: is calculated with classifier-generated labels (generated by inputing the original image into four commonly used classifiers, since each classifier can be regarded as a DMV systerm due its human- comparable performance) instead of their corresponding human-annotated label. The reason for using classifier-generated labels instead of human-annotated one is for the DMV-JND modeling requirement elaborated in Subsection \ref{III-A}.

With massive data and high-performance GPU hardware, Deep Machine Vision (DMV) has made breakthroughs in many machine vision tasks, such as image classification \cite{wang2017residual}, object detection \cite{lin2017feature}, person re-identification \cite{zhang2021hybrid}, and so on. It also makes the ultimate receiver and appreciator of increasingly larger number of images and videos change from the HVS to the DMV. Many images and videos processing applications are developed for the DMV now, and we naturally wonder: does the DMV have the JND? \textcolor{black}{Unlike the HVS-JND aiming to find the visual redundancy for the HVS, the JND for the DMV is to find the redundancy of images and videos for deep machine vision by considering the effects of such redundancy during the DMV tasks. If the DMV has JND}, the JND for the DMV will greatly benefit the DMV-oriented visual computing applications. %For instance, it would help to design novel codecs for AI-oriented image/video compression \cite{chen2019toward} \textcolor{black}{via a DMV-JND inspired bit allocation strategy, e.g., lower/higher bit is assigned to the region with more/lower redundancy for DMV to achieve overall bit saving within the tolerance of the DMV.} 
% For instance, it would help to design novel codecs for DMV-oriented image and video compression \cite{chen2019toward} \textcolor{black}{via a DMV-JND inspired bit allocation strategy, e.g., a lower/higher bit is assigned to a pixel with more/lower redundancy for DMV to achieve overall bit saving within the tolerance of a DMV task.}
For instance, it would help to design novel codecs for DMV-oriented image and video compression \cite{chen2019toward} \textcolor{black}{via a DMV-JND inspired bit allocation strategy. For example, the lower bit is assigned to pixels with higher redundancy for the DMV, while the higher bit is assigned to pixels with lower redundancy so as to achieve overall bit saving.} Besides, it may provide us a novel perspective for a wider scope, e.g., DMV-oriented quality evaluation for natural images/videos, computer-generated graphics/animation \cite{wang2018video}, style transformation images/videos \cite{yi2017dualgan}, and even rethinking of the deep neural network security (e.g., adversarial attack \cite{goodfellow2014explaining, moosavi2016deepfool}). More details on DMV-oriented potential applications will be highlighted in Section \ref{2}. 

In this paper, we make an initial exploration on the JND for the DMV, and propose the first model to demonstrate the existence of DMV-oriented JND, termed as the DMV-JND. As shown in Fig. \ref{RCA}, our generated DMV-JND can be tolerated by the image classification task, while the White Gaussian Noise (WGN) with the same amount of noise will be
%the same level random noise (White Gaussian Noise, WGN) is 
noticed by the DMV and lead to significantly lower \textbf{Relative Classification Accuracy (RCA)}. RCA is calculated with classifier-generated labels instead of their corresponding human-annotated ones (i.e., generated by inputting the original image into four commonly used classifiers, as to be discussed in Subsection \ref{III-A}). %The reason for using classifier-generated labels instead of the human-annotated ones is to be elaborated in Subsection \ref{III-A}.

The main contributions in this research are summarized as follows. 
\begin{itemize}
	\item To the best of our knowledge, our work is the first to demonstrate that the DMV has the JND. Besides, we also propose the first algorithmic framework to model the DMV-JND.
%	\item The proposed DMV-JND model, achieved via a DMV-JND-NET with unsupervised learning, is capable of tolerating distorted images with average PSNR of only 9.56dB in DMV, which is quite smaller than the 25 to 35 dB in HVS. This demonstrates that DMV can tolerate more changes than HVS.
%	\item DMV is capable of tolerating JND-distorted image with average PSNR of only 9.56dB, generated via our DMV-JND model by using DMV-JND-NET with unsupervised learning. This quite smaller than the 25 to 35 dB that HVS can tolerated. Therefore, compared with HVS, DMV can tolerate more changes.
	\item The proposed DMV-JND model, achieved via unsupervised learning with our DMV-JND-NET, is capable of generating the DMV-JND distorted image with average PSNR of only 9.56dB for the DMV. %which is significantly smaller than the scale of 25 to 35 dB for the HVS. This demonstrates that DMV can tolerate more changes than the HVS.
	\item A semantic-guided redundancy assessment strategy is introduced toward the reasonability of the generated DMV-JND, by restraining its magnitude and spatial distribution.
	%	\item To ensure the generated JND can be tolerated by DMV, we utilize the cross-entropy loss between the original image and the JND-distorted image. However, such restrain is still limited to the JND generation. We further propose a semantic-guided redundancy assessment strategy for different visual content by leveraging CAM maps, which is integrated into the DMV-JND-NET as JND-magnitude loss and JND-spatial-distribution loss to control the magnitude and spatial distribution of JND generation elegantly. 
	%	\item Our work can make the DMV tolerate 9dB noise without its classification accuracy, which can be used to guide the DMV-oriented image/video coding. 
	\item Reducing the noise from the resultant DMV-JND to zero, the DMV maintains the RCA throughout the process; This demonstrates that the DMV has the same homogeneous property as the HVS-JND. %It makes the proposed DMV-JND model can be used in DMV-oriented image/video processing. 
	
\end{itemize}

\section{Related Work}
\label{2}
In this section, the HVS-JND and its applications are first reviewed. %After that, the significance of DMV-JND and its potential applications are introduced.% 
Then, to distinguish the DMV-JND from adversarial attack, the review of adversarial attack is elaborated. Finally, CAM \cite{zhou2016learning}, a technique that exposes the implicit attention of CNNs absorbed in our proposed DMV-JND, is briefly reviewed.

\subsection{Techniques and applications of the HVS-JND}
\label{II-A}
There have been substantial researches in the HVS-JND during the past decades \cite{jayant1993signal,chou1995perceptually,yang2005just,liu2010just,wei2009spatio,bae2013novel,jin2016statistical,wu2017enhanced,wang2017videoset,liu2019deep,zhang2021deep}. Commonly, the HVS can tolerate distorted images with average PSNR from 25 to 35 dB for the existing HVS-JND models \cite{chou1995perceptually,yang2005just,liu2010just,wu2017enhanced}. These models can be divided into two categories: i) pixel-domain-HVS-JND models that can directly obtain the HVS-JND threshold of each pixel by leveraging the background luminance and various textural masking effects \cite{chou1995perceptually,yang2005just,liu2010just}; ii) sub-band-domain-HVS-JND models that usually transfer the pixel domain image to the sub-band domain one via discrete cosine transformation (or the other transformation) first, and then the HVS-JND threshold on sub-band domain is estimated by taking the contrast sensitivity function, luminance adaptation \cite{bae2013novel}, contrast masking \cite{wei2009spatio}, and textural masking effects into account. However, the HVS-JND threshold for each pixel or sub-band is often separately estimated in the pixel or sub-band domain and summed up in a local neighborhood, which is not capable of representing the total masking of the whole image. Some recent works \cite{jin2016statistical,wang2017videoset,liu2019deep,zhang2021deep} propose learning-based HVS-JND models using databases with the subjective HVS-JND tests, which are conducted at the whole picture or video frame level.

The HVS-JND is able to predict the visual redundancy of images and videos, which makes it widely used in many image and video processing related applications, such as the aforementioned HVS oriented perceptual visual signal compression, watermarking, image and video quality evaluation, and so on. The goal of the HVS oriented perceptual visual signal compression is to achieve bit saving while maintaining good perceptual quality. To this end, the pixels with higher redundancy tolerating more noise without perceived by the HVS (without sacrificing perceptual quality) are compressed with lower bits, while the pixels with lower redundancy tolerating less noise are compressed with higher bits.
%A pixel with more/lower redundancy tolerates more/less noise without perceived by the HVS (without sacrificing perceptual quality), and can compressed with lower/higher bits to achieve bit saving. 
Therefore, the HVS-JND can be used to guide the bit allocation \cite{luo2013h,zhou2020just} during rate distortion optimization and motion estimation speeding up \cite{yang2005just}. Besides, the HVS-JND can be used to design filters \cite{kim2015hevc} to reduce the redundancy information and further achieve bit saving during the filtered image compression. For the HVS-JND oriented watermarking applications \cite{chou2010perceptually}, the pixels with certain redundancy will be added with well-handcrafted noise (i.e., watermarking) before transmission. As the added noise cannot be perceived by the HVS (under the HVS-JND), which ensures the information hided in the noise secure. Only the person with the codebook at the receiver side can recover the hidden information successfully. During the HVS-JND oriented video quality evaluation \cite{zhang2018deepqoe}, under the same level of distortion, there will be unobvious quality degradation on the pixel with higher redundancy (lower sensitivity). However, the obvious quality degradation will appear on the pixel with lower redundancy (higher sensitivity).

%, which is the main idea used in HVS-JND oriented image/video quality evaluation. 

All the techniques and advantages of the HVS-JND above can only be applied to the applications in which the HVS is the ultimate receiver and appreciator. However, increasingly larger number of images and videos are used to perform the DMV tasks instead of being viewed by humans. To better process data and optimize applications for the DMV, exploration on the redundancy of image and video for the DMV (i.e., DMV-JND) is highly significant. Thus, the JND for the DMV should be studied and designed.

\subsection{Adversarial Attack}
\label{2C}
As deep neural networks (DNNs) have achieved great successes in many applications, the security of DNNs has attracted more attention in recent years. Adversarial attacks are used to craft adversarial samples \cite{goodfellow2014explaining} to fool the DNNs, which are expected to have similar appearances with the clean ones, while can produce high-confidence incorrect predictions. Generally, adversarial attacks can be divided into two categories, i.e., white-box attacks \cite{kurakin2016adversarial,moosavi2016deepfool}
\begin{figure}[htbp]
	\begin{center}
		\noindent
		\includegraphics[width = 3.4 in]{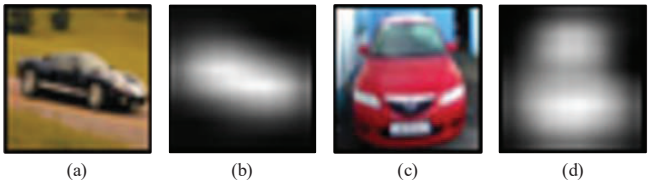}
		\caption{Illustration of the important semantics. (a) and (c) are the original images. Their associated CAM maps are (b) and (d).}\label{Car2}
	\end{center}
\end{figure} 
and black-box attacks \cite{cheng2019improving,ilyas2018prior,yan2019subspace,papernot2017practical,moosavi2017universal,dong2019evading}. White-box attacks are easier to fool the DNNs with their complete information, e.g., the structure and parameters of the victim DNNs, which can find a more effective way to attack. For the generalized application scenes, the complete information of various DNNs is hardly obtained before attacks. Black-box attacks are more applicable to various DNNs, which need to attack their common vulnerability. The ultimate positive goal of the related research is to build more robust DNNs based on the vulnerabilities, which are expected to be able to effectively defense more kinds of attacks.

\subsection{Class Activation Mapping (CAM)}
\label{2D}
CAM was first proposed by Zhou $et\ al.$ \cite{zhou2016learning}. They found that revisiting the global average pooling layer for image classification could actually build a localizable representation to expose the implicit attention of CNNs on an image. By utilizing this technique, the importance of different regions in the image, which leads to the image being classified to the specified class, can be represented with a CAM map. As shown in Fig. \ref{Car2}, when original image (a) and (c) are classified as ``$car$'', the important regions corresponding to ``$car$'' will be highlighted in their associated CAM maps (b) and (d). The value of each pixel in the CAM map is from 0 to 1. The brightness of the pixel indicates the importance of pixel for the specified class during image classification task. Given the specified class, CAM will locate its corresponding object in the image. Hence, it is widely used in many weakly supervised object localization works \cite{zhang2018top,wei2017object,zhang2018self,zhang2018adversarial}.

\section{Description, Formulation, and Potential Applications of the DMV-JND}
\label{S-II}
\subsection{Description}
\label{III-A}
The HVS-JND models aim to find the visual redundancy of images or videos, by finding the HVS-JND (threshold) of each pixel (or the associated sub-bands), as already introduced in Subsection \ref{II-A}. Any changes on each pixel under its associated HVS-JND cannot be perceived by the HVS. A benchmark of the HVS-JND modeling assessment is the magnitude of the tolerated HVS-JND: without being perceived by the HVS, the higher HVS-JND is tolerated, the better HVS-JND model is. Similarly, the DMV-JND model is defined in this paper with reference to the HVS-JND one. Unlike human eyes being the final receptor of the HVS-JND, our proposed DMV-JND model is the DMV task oriented. Therefore, our proposed DMV-JND model is to find the visual redundancy characteristic of the images or videos for the DMV task for each pixel. Any changes under the DMV-JND magnitude will not affect the performance of the DMV task. We select the image classification as the DMV task in this paper, and the extension to other types of the DMV tasks will be demonstrated in future work.     

As mentioned above, the HVS-JND models are built for human eyes. The difference between the HVS-JND-distorted image and the original one is commonly assessed by human subjects. The assessment results are taken as the ground truth during building the HVS-JND data set \cite{jin2016statistical,wang2017videoset}. In this paper, classifier-generated labels are refer to the labels generated with four commonly used classifiers (AlexNet \cite{krizhevsky2012imagenet}, VGG \cite{simonyan2014very}, ResNet \cite{he2016deep}, and DenseNet \cite{huang2017densely}), which are chosen as our ground truth labels, since the final receptors in our model are the DMV (classifiers). Each of these classifiers can be regarded as a successful artificial machine vision system due to its high performance. Although there are small differences among them, the main techniques (convolution, pooling, randomly dropout and so on) that they used are the same, which suggests that they may have similar DMV-JND characteristics. Therefore, selecting classifier-generated labels as ground truth instead of the human-annotated ones (generated by human eyes) is more reasonable and generalizable for our DMV-JND modeling.

\subsection{Formulation}
\label{III-B}

Assume that $C_n(\cdot)$ is the $n^{th}$ classifier in these four classifiers, where $n = 1, 2, 3, 4$. $x$ denotes the original image. Its associated DMV-JND image and the DMV-JND distorted image are $e$ and $\hat{x}$, respectively; $\hat{x} = x + e$. Therefore, the classifier-generated labels (regarded as ground truth labels) and their associated distorted ones, generated by feeding $x$ and $\hat{x}$ to the $n^{th}$ classifier, can be denoted by $C_n(x)$ and $C_n(x+e)$. Then, the DMV-JND modeling can be initially formulated as 
\begin{equation}
\label{init}
\arg \min _{e}\left(\sum_{n} CE\left(C_{n}(x+e), C_{n}(x)\right)+ \gamma \cdot \frac{1}{|e|}\right),
\end{equation}
where $CE(\cdot)$ denotes the cross-entropy loss. The first term in Eq. \eqref{init} ensures the distortion in the DMV-JND distorted image can be tolerated by these four classifiers. The second term requires the DMV-JND $e$ as large as possible.  \textcolor{black}{$\gamma$ is a weight to balance these two items.} Therefore, the DMV-JND modeling can be achieved by learning a reasonable DMV-JND image $e$ in a network. To achieve this, we propose a DMV-JND-NET. As no human-annotated labels are used during training (four classifier-generated labels are used as the ground truth ones), our proposed DMV-JND-NET can generate the DMV-JND with unsupervised learning. However, during our exploration, the restraint on the DMV-JND generation above is still not sufficient, which cannot well control the DMV-JND generation, especially for training process. The results of the generated DMV-JND are not regulated.

The HVS-JND models commonly have a redundancy assessment strategy, which makes the HVS-JND model adaptively adjust the magnitude and the spatial distribution of HVS-JND according to the content of image: 
\begin{figure*}[htbp]
	\begin{center}
		\noindent
		\includegraphics[width = 6.3in]{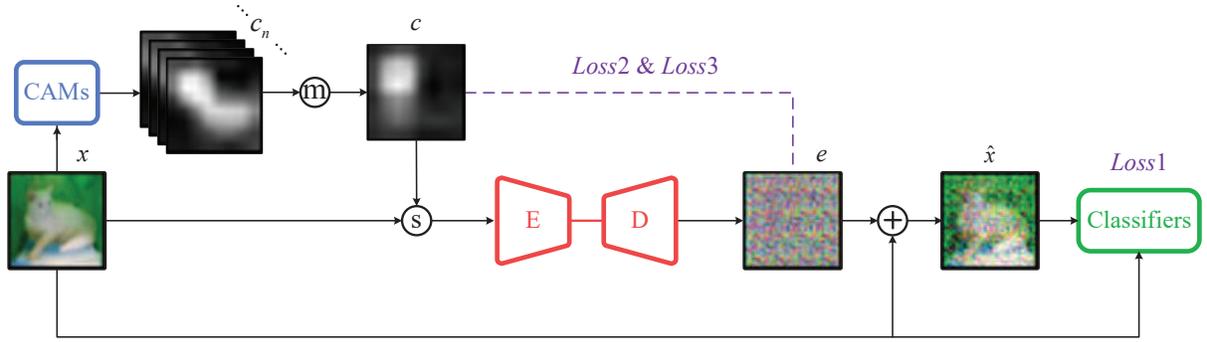}
		\caption{The framework of the proposed DMV-JND-NET. First, attention maps $c_n$ of the original image $x$ are generated via CAM techniques \cite{zhou2016learning} and merged into $c$. Then, $c$ together with $x$ are stacked and fed into E-D to generate the DMV-JND $e$. To well restrain the magnitude and spatial distribution of $e$, a semantic-guided redundancy assessment strategy is proposed and formulated as $Loss2$ and $Loss3$. Besides, $Loss1$, the cross-entropy loss between $x$ and its associated DMV-JND distorted image $\hat{x}$ is used to guarantee that the DMV-JND will not change the classification performance.}\label{Flowchart}
	\end{center}
\end{figure*}
insensitive regions tolerate more noise, given with larger thresholds, while smaller thresholds are assigned to the sensitive regions. 

To better control the DMV-JND generation, a similar redundancy assessment strategy is proposed in this paper, termed as semantic-guided redundancy assessment strategy. In this paper, our target is to achieve image classification with the DMV. Hence, the sensitivity of the DMV can be represented by the attention of classifier. As shown in Fig. \ref{Car2} (a), there are cars, trees, and ground. Assume that the class is specified ``$car$''. The DMV focuses on the pixels at the car regions due to its high related semantic ``$car$''. Although the pixels located at trees and ground contain some other semantics, they are ignored due to their unrelated semantics. Therefore, semantic-guided redundancy assessment strategy can be summarized as: the pixel with high related semantic has lower redundancy, which tolerates less noise, assigned with smaller DMV-JND, while a larger DMV-JND is assigned to the pixel with larger redundancy due to its unrelated semantics. Since CAM map well reflects the semantic relevance of different pixels during image classification task, it is used as a reference in our semantic-guided redundancy assessment strategy. Moreover, the proposed assessment mechanism is further utilized to design two sub-losses (magnitude loss and spatial distribution loss), which restrains the magnitude and spatial distribution of the DMV-JND generation during training, respectively. More details will be elaborated in Section \ref{S-III}.   

\subsection{Potential Applications}
\label{III-C}
Once the DMV-JND is obtained, its potential applications would be greatly attractive, e.g., the DMV-oriented image and video codecs. As the DMV-JND can predict the redundancy of the image for the DMV, more redundant pixels are compressed with larger quantization parameter (QP), while less redundant pixels are compressed with smaller QP. According to our experiences, the quantization step is set as twice of the DMV-JND during bit allocation to guarantee that the quantization-caused errors are still tolerated by the DMV to avoid the performance decreasing of the DMV task. In this case, we will achieve bits saving, meanwhile maintains the same task performance, e.g., the classification accuracy, detection accuracy, and so on. Besides, the application of watermarking is also based on the redundancy of the image or video as aforementioned, but the redundancy here is for the DMV instead of the HVS. The basic theory of the DMV-JND oriented watermarking is similar to that of the HVS-JND oriented watermarking. So is the image and video quality assessment for the DMV. Moreover, the DMV-JND will also bring the rethinking of the deep neural network security, e.g., adversarial attacks. 

It should be noticed that our DMV-JND is quite different from adversarial attacks. First, there is a different goal: adversarial attacks are to find the most effective attack way (e.g., minimal noise) to change the final result (e.g., the label in the image classification task). By contrast, the DMV-JND tries to maintain the original result of the network by adding noise as much as possible. Second, there are different applications: adversarial attacks aim to find the vulnerability of current networks, thus further building a robust network, while our work is to find the redundancy of the DMV for the basic DMV-oriented image and video processing (e.g., compression and watermarking). Besides, adversarial attack is more likely to find an optimal perturbation value, which may not guarantee that all the changes below it will make the network achieve the same result. However, our work is to find the DMV-JND threshold (boundary), and any changes under the DMV-JND will be tolerated by the DMV, i.e., maintaining the same result of networks as explored and exhibited in Subsection \ref{IV-F}, which will make it widely used in the DMV-oriented image and video processing.    

\section{DMV-JND-NET}
\label{S-III}
\subsection{Architecture}  
\label{architecture}
The major components of the DMV-JND-NET include: three network parts (CAMs, E-D, Classifiers), and three operations ($\textcircled{m}$, $\textcircled{s}$, $\textcircled{+}$), as shown in Fig. \ref{Flowchart}. Specifically, $\textcircled{m}$ (merging operation) merges several images into one by applying a weighting calculation on corresponding pixels. $\textcircled{s}$ (stacking operation) stacks images together. $\textcircled{+}$ (element-wise addition) adds one image to another by pixel-wise addition. The details will be elaborated in Subsection \ref{architecture}.

\begin{figure}[htbp]
	\begin{center}
		\noindent
		\includegraphics[width = 3.4 in]{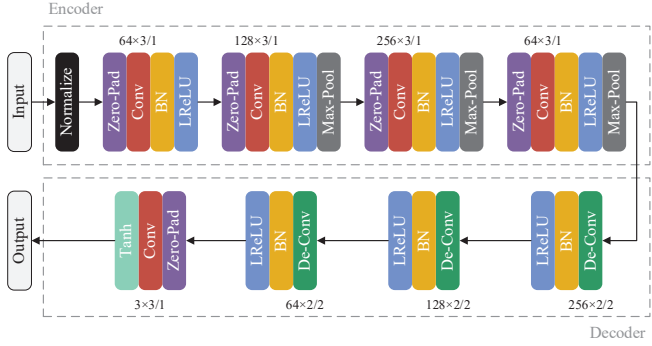}
		\caption{Illustration of the hybrid convolutional autoencoder architecture used in this work. The notation $A \times B$ refers to $B \times B$ convolutions with $A$ filters. The number following the slash indicates stride in the case of convolutions. The zero padding and max pooling used here are set to $(1,1,1,1)$ and $(2,2)$, respectively. BN is the short form for batch normalization.}\label{AE}
	\end{center}
\end{figure}

CAMs are used to generate a merged CAM map. On the one hand, the merged CAM map is used to help to generate much richer features for the DMV-JND generation. On the other hand, it is used as a reference during the semantic-guided redundancy assessment of each input image (sample). CAMs is made up of four CAM networks, which are revised from the aforementioned four classifiers (AlexNet, VGG, ResNet, and DenseNet) that appear as Classifiers in the Fig. \ref{Flowchart}. The revising and setting of CAMs in this paper refer to \cite{zhou2016learning}. These four CAM networks have been pre-trained, and their weights are not updated during the DMV-JND-NET training.

E-D (hybrid convolutional autoencoder) is used to generate the DMV-JND according to the input image and its associated merged CAM map. It mainly contains a convolution plus pooling structure encoder and a large stride deconvolution structure decoder. A convolution plus pooling structure encoder is good at extracting robust features, which is widely used in image denoising \cite{chaitanya2017interactive}. A large stride deconvolution structure decoder can generate details, which is commonly used in image super-resolution \cite{cheng2018performance} and restoration \cite{mao2016image}. With such a hybrid convolutional autoencoder, the semantic features can be robustly extracted at the encoder. Meanwhile, more elaborated JND can be generated at the decoder side according to the extracted semantic features. Besides, padding and normalization operations are also used in E-D. A more detailed structure of E-D refers to Fig. \ref{AE}. The parameters of this network are randomly initialized at first and then updated during training.

Classifiers are used to generate the classifier-generated labels and their associated distorted labels by feeding the original image and its associated DMV-JND distorted one. Such two kinds of labels are further used to calculate a cross-entropy loss in the loss function. Similarly, all the classifiers are pre-trained, and their weights are not updated during the DMV-JND-NET training.

Assume that the original image $x$ is fed to the well trained CAMs at first. CAMs is made up of four CAM networks, denoted by $C_1(\cdot)$, $C_2(\cdot)$, $C_3(\cdot)$ and $C_4(\cdot)$. Then, four CAM maps are generated, denoted by $c_1$, $c_2$, $c_3$, and $c_4$. For the $n^{th}$ CAM map, we have $c_n = C_n(x)$, where $n=1,2,3,4$. After that, the merging operation is applied to four CAM maps and a merged CAM map is generated, denoted by $c$. We have
\begin{equation}
\label{c}
c=\frac{1}{4} \sum_{n}  c_{n}.
\end{equation}	

Then, the merged CAM map $c$ together with the original image $x$ are stacked via stacking operation and fed to the hybrid convolutional autoencoder E-D, denoted by $E(\cdot)$. And then, the DMV-JND image is generated under the loss function to be explored in the next subsection, which is denoted by $e$. We have 
\begin{equation}
\label{c}
e = E(x, c).
\end{equation}

After that, the generated DMV-JND image $e$ is added to the original image $x$ by applying element-wise add, and we obtain the DMV-JND distorted image $\hat{x}$, as we have already presented at the start of Subsection \ref{III-B}:
\begin{equation}
\label{c}
\hat{x} = e + x.
\end{equation}  

%\begin{figure}[htbp]
%	\begin{center}
%		\noindent
%		\includegraphics[width = 2.0in]{Car2.eps}
%		\caption{The important semantic comparison between two images with the same class label "car". (a) and (c) are the two original images. Their associated CAM maps are (b) and (d). The important semantic in (a) and (c) can be represented by the brightness of (b) and (d). Hence, (a) has larger important semantic.}\label{Car2}
%	\end{center}
%\end{figure}

Furthermore, the DMV-JND distorted image $\hat{x}$ and original image $x$ are fed to Classifiers, which contains four corresponding classifiers, denoted by $S_1(\cdot)$, $S_2(\cdot)$, $S_3(\cdot)$ and $S_4(\cdot)$. For the $n^{th}$ classifier, the distorted and original softmax values are denoted by $\boldsymbol{\hat{s}_n}$ and $\boldsymbol{s_n}$, we have 
\begin{equation}
\left\{\begin{array}{l}
\boldsymbol{\hat{s}_{n}}=S_{n}(\hat{x}) \\
\boldsymbol{s_{n}}=S_{n}(x).
\end{array}\right.
\end{equation}
It should be noticed that $\boldsymbol{\hat{s}_n}$ and $\boldsymbol{s_n}$ are two vectors of probabilities of different labels. Assign the labels by finding the indexes of the biggest elements in $\boldsymbol{\hat{s}_n}$ and $\boldsymbol{s_n}$, and this process is represented by function $M(\cdot)$:
\begin{equation}
\left\{\begin{array}{l}
\hat{l}_{n}=M\left(\boldsymbol{\hat{s}_{n}}\right) \\
l_{n}=M\left(\boldsymbol{s_{n}}\right),
\end{array}\right.
\end{equation}  
where $\hat{l}_{n}$ and $l_{n}$ are the distorted label and classifier-generated label with the $n^{th}$ classifier, respectively. 

\subsection{Loss Functions}
The overall loss function is made up of three sub-losses, namely cross-entropy loss, magnitude loss, and spatial distribution loss, denoted by $Loss1$, $Loss2$, and $Loss3$, receptively. They have different functionalities during generating the DMV-JND. The overall loss is denoted by $Loss$, we have
\begin{equation}
\label{alpha}
Loss=Loss1 + \alpha \cdot Loss2 + \beta \cdot Loss3,
\end{equation}
where $\alpha$ and $\beta$ are two weights to balance these three sub-losses, and their settings refer to Subsection \ref{Ex_set}. 

% The cross-entropy loss between the DMV-JND distorted image $\hat{x}$ and original image $x$ is to guarantee that the generated DMV-JND will be tolerated by all these four classifiers during image classification. 
The cross-entropy loss between the DMV-JND distorted image $\hat{x}$ and original image $x$ is to guarantee that the generated DMV-JND will be tolerated by all these four classifiers. 
For the $n^{th}$ classifier, we have
\begin{equation}
Loss1_{n}=CE(\boldsymbol{\hat{s}_{n}}, l_{n}). 
%= \text{cross-entropy}(S_{n}(\hat{x}), M(S_{n}(x)))
\end{equation}
Therefore, $Loss1$ can be represented as
\begin{equation}
Loss 1=\frac{1}{4} \sum_{n} Loss1_{n}.
\end{equation}

For different images, the magnitudes of redundant semantic are not the same, even though they may be specified with the same class label. As shown in Fig. \ref{Car2}, the dark region of (b) is larger than that of (d), and this means that there is higher redundancy in (a) compared with (c). Our semantic-guided redundancy assessment strategy indicates that (a) is assigned with the larger DMV-JND. To restrain the magnitude of the DMV-JND according to the content of the original image $x$, magnitude loss is introduced. The average magnitude of the 
high related semantic of each pixel in $x$ is denoted by $I$, which can be defined as follows, based upon its associated merged CAM map $c$:
\begin{equation}
I=\frac{1}{H W} \sum_{h=1}^{H} \sum_{w=1}^{W} c(h, w).
\end{equation}
$H$ and $W$ are the height and width of the image $x$ and $c$. $c(h,w)$ ($c(h,w)\in [0,1]$) is the value of pixel $(h,w)$ in $c$. Therefore, the average magnitude of the unrelated semantic (redundancy) of each pixel, namely the targeted average DMV-JND, denoted by $N$, can be defined as
\begin{equation}
N = 1 - I.
\end{equation}

The actual averaged DMV-JND generated with our DMV-JND-NET is denoted by $N_0$, we have
\begin{equation}
N_{0}=\frac{1}{H W} \sum_{h=1}^{H} \sum_{w=1}^{W}|e(h, w)|,
\end{equation}
where $|\cdot|$ is the operation of taking absolute value. Then, $Loss2$ can be formulated as
\begin{equation}
\begin{aligned}
Loss2 &=\ln \left(\frac{N^{2}+N_{0}^{2}+q}{2 N N_{0}+q}\right) \\
&=\ln \left(N^{2}+N_{0}^{2}+q\right)-\ln \left(2 N N_{0}+q\right),
\end{aligned}
\end{equation}
where $q$ is a small constant (set to $1^{-10}$) to prevent the denominator from being zero.

%\begin{equation}
%\left\{\begin{array}{l}
%T=\frac{1}{N+q} \\
%T_{0}=\frac{1}{N_{0}+q},
%\end{array}\right.
%\end{equation} 
%where $q$ is a constant to prevent the denominator from being zero, set to $1e-10$. Then, $Loss2$ can be formulated as
%\begin{equation}
%\begin{aligned}
%Loss2 &=\ln \left(\frac{T^{2}+T_{0}^{2}+q}{2 T T_{0}+q}\right) \\
%&=\ln \left(T^{2}+T_{0}^{2}+q\right)-\ln \left(2 T T_{0}+q\right).
%\end{aligned}
%\end{equation}

Initially, $N$ is smaller than $N_0$. $Loss2$ makes $N$ approach to $N_0$ during training. It makes sure that the actual average DMV-JND will be increasingly generated during training. 

% After the magnitude control of generated DMV-JND with magnitude loss above, the spatial distribution of DMV-JND should also be restrained according to the semantic-guided redundancy assessment strategy: a pixel with lower redundancy is assigned with a smaller DMV-JND, while the one with higher redundancy is assigned with a larger DMV-JND. 
After the magnitude control of the generated DMV-JND above, the spatial distribution of the DMV-JND should also be restrained via the semantic-guided redundancy assessment strategy. That is, a pixel with lower redundancy is assigned with a smaller DMV-JND, while the one with higher redundancy is assigned with a larger DMV-JND. 
To achieve this, the merged CAM map $c$ is utilized again. Firstly, a vector $\boldsymbol{v}$ is generated by applying softmax operation on $c$, and we have
\begin{equation}
\boldsymbol{v}=\text{softmax}(c).
\end{equation}
% Then, apply inner product $\left \langle \cdot, \cdot \right \rangle$ between $\boldsymbol{v}$ and $e$. The spatial distribution loss is formulated as
Then, by employing inner product $\left \langle \cdot, \cdot \right \rangle$ between $\boldsymbol{v}$ and $e$, the spatial distribution loss is formulated as
\begin{equation}
\label{loss3}
Loss3=\left \langle \boldsymbol{v}, e \right \rangle.
\end{equation}
$Loss3$ makes sure that larger DMV-JND is generated on the pixels with higher redundancy. 

\begin{figure}[htbp]
	\begin{center}
		\noindent
		\includegraphics[width = 2.0in]{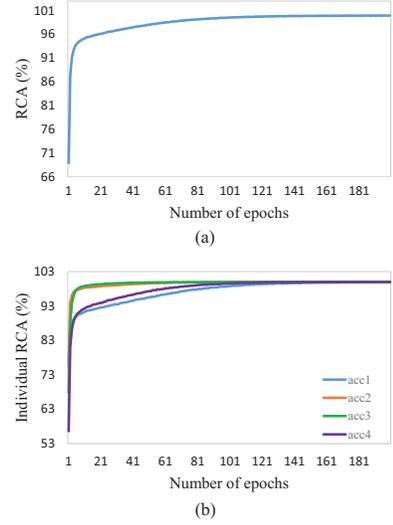}
		\caption{Training RCA trend, (a) is the RCA trend. Its associated individual RCA trend for each classifier is exhibited in (b).}\label{Acc2}
	\end{center}
\end{figure}

\section{Experiments}
\subsection{Dataset and Settings}
\label{Ex_set}
We evaluate the proposed DMV-JND-NET on the CIFAR-10 dataset \cite{krizhevsky2009learning}, which is widely used benchmark for image classification. The CIFAR-10 dataset consists of 60,000 32x32 color images in 10 classes, with 6,000 images per class, which are divided into 50,000 training images and 10,000 test images. All the classifiers used, including AlexNet, VGG, ResNet, and DenseNet, are pre-trained. The CAM nets used in this paper, such as AlexNet-CAM, VGG-CAM, ResNet-CAM, and DenseNet-CAM, are with the settings used in paper \cite{zhou2016learning}. All the experiments are conducted on one NVIDIA GTX 1080 GPU with 8GB memory. 

During the training of the DMV-JND-NET, each sample from CIFAR-10 is flipped horizontally with a probability of 0.5 and normalized with normalization parameter ((0.5,0.5,0.5),(0.5,0.5,0.5)). The Adam optimizer with a batch size of 50 examples, learning rate of 1e-5, and weight decay of 1e-3 is adopted here to minimize training loss. $\alpha$ and $\beta$ in Eq. \eqref{alpha} are set to 1 in this work.

\subsection{RCA and Loss}
\label{IV-B}
The formulation of RCA is given in this subsection first. For the $n^{th}$ classifier, its individual RCA is represented as
\begin{equation}
a c c_{n}=\frac{100 \%}{P} \sum ^{P} \text{pre}\left(\hat{l}_{n}, l_{n}\right).
\end{equation}
$\sum\text{pre}(\cdot)$ is used to sum all $P$ prediction results, where the right and false results are represented as 1 and 0, respectively. Then, RCA is formulated by the average of four individual RCAs
\begin{equation}
a c c=\frac{1}{4} \sum_{n} a c c_{n}.
\end{equation}

Fig. \ref{Acc2} (a) shows the trend of $acc$. It contains %
\begin{figure*}[htbp]
	\begin{center}
		\noindent
		\includegraphics[width = 7.15in]{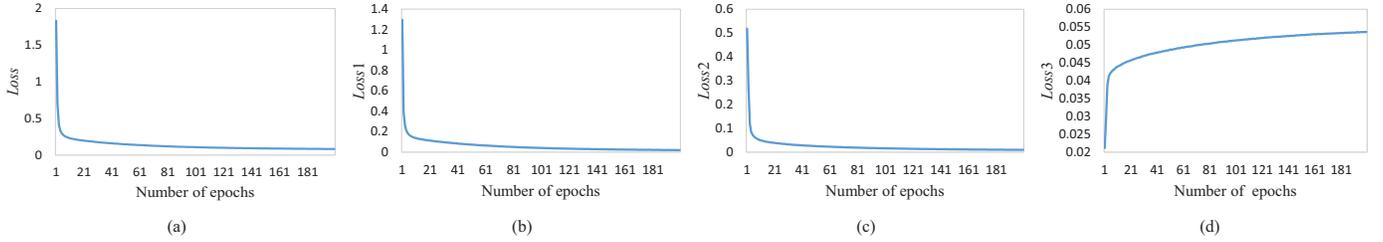}
		\caption{Training loss trend. (a) is the $Loss$ in different epochs. Its associated $Loss1$, $Loss2$, and $Loss3$ (notice the vertical scale difference) are exhibited in (b), (c), and (d), respectively.}\label{Loss2}
	\end{center}
\end{figure*}
\begin{figure}[htbp]
	\begin{center}
		\noindent
		\includegraphics[width = 2.0in]{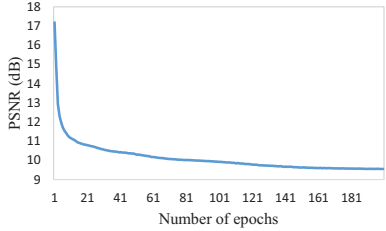}
		\caption{Trend of the PSNR between original image and its associated DMV-JND distorted one.}\label{MSE2}
	\end{center}
% 	\vspace{-0.5cm}
\end{figure}
three stages: i) the fast process of RCA increasing stage is about from epoch 1 to epoch 10; ii) the slightly increasing stage is about from epoch 11 to epoch 140; iii) the stable stage is about from epoch 141 to epoch 200 when RCA reaches 100\%. The result means that even the DMV-JND is added via our DMV-JND model, the DMV-JND distorted images still can be correctly classified by all these four classifiers. Besides, the individual RCA trend of each single classifier ($acc_1$, ..., $acc_4$) reaches 100\% as well, as shown in Fig. \ref{Acc2} (b). It should be mentioned that all the results above are for training RCA. The testing RCA of the proposed DMV-JND-NET reaches 92.18\%, and its associated individual RCA reaches 91.17\%, 93.03\%, 93.48\%, and 91.05\%, respectively. Although we cannot achieve a better testing RCA under such a magnitude of the DMV-JND, it is enough to demonstrate that our generated DMV-JND can be tolerated by four commonly used classifiers. We believe this problem will be better solved in the following research. 

Conversely, as epoch increases, the overall loss drastically decreases at first. Then, it's falling at a slower pace and finally converges to a value of 0.083, as shown in Fig. \ref{Loss2} (a). In addition, the trends of $Loss1$, $Loss2$, and $Loss3$ are also respectively shown in Fig. \ref{Loss2} (b), (c), and (d). $Loss1$ and $Loss2$ have a falling trend. However, the trend of $Loss3$ is an increasing one. That's because as the training epoch increases, $Loss2$ decreases and $N_0$ approaches $N$. Therefore, larger and larger DMV-JND is generated. It leads to the increase of $e$. Then, $Loss3$ increases. However, $Loss3$ still restrains the spatial distribution of the DMV-JND. All the demonstrates above are further supported by the experimental results in Subsections \ref{IV-C} and \ref{IV-D}.

%In this work, compared with $Loss3$, $Loss2$ is the dominant loss due to its larger value, which makes sure the amount of JND increases. However, $Loss3$ still works as a reference loss, which still restrains the location of JND. This conclusion will be further proved in Section \ref{IV-C} and \ref{IV-D}.

\subsection{PSNR Measure}
\label{IV-C}
Fig. \ref{MSE2} shows the trend of the Peak Signal-to-Noise Ratio (PSNR) between the original image $x$ and the DMV-JND distorted image $\hat{x}$. As the epoch number increases, PSNR firstly decreases dramatically. Then, the decreasing trend becomes a gradual process and PSNR eventually converges to 9.56 dB. The result is in line with our discussion in Subsection \ref{IV-B}: as the number of epoch increases, larger DMV-JND is generated with the decreasing of $Loss2$, which leads to the increasing tendency of PSNR between the original image and the DMV-JND distorted image.

However, as the number of epoch increases, RCA increases as well. It means that although larger DMV-JND is generated via DMV-JND-NET, the spatial distribution of the generated DMV-JND becomes more reasonable; That's because the generated DMV-JND is adjusted to more reasonable regions under the control of $Loss3$; In other words, the DMV-JND-NET is optimized towards the right direction. More visual evaluation results will be exhibited in the next subsection to support this point. It is worth to mention that the DMV can tolerate the DMV-JND distorted image with about 9.56 dB in our work, which is significantly smaller than the 25 to 35 dB in the HVS as mentioned in the Related Work (Subsection \ref{II-A}). This opens a new horizon for visual feature compression toward the DMV. Of course, it needs to be understood that the HVS-JND is an absolute JND for situations where the difference is undetectable in a strict psychophysical sense, while the DMV-JND is a utility-oriented JND that merely does not affect the intended utility (image classification in this work). More research is called for investigation for their differences \cite{Lin20c}.

\subsection{Visual Evaluation}
\label{IV-D}
\begin{figure*}[htbp]
	\begin{center}
		\noindent
		\includegraphics[width = 7.15in]{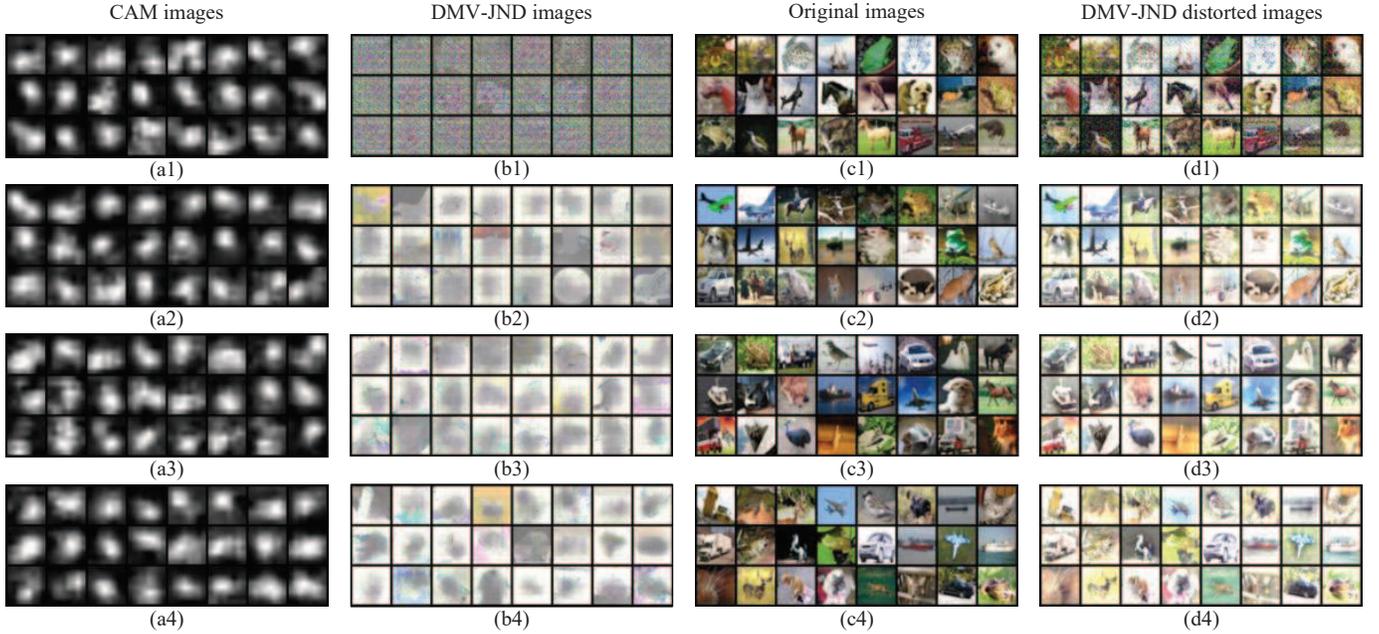}
		\caption{(a1)-(a4) are the CAM maps in epoch 1, 15, 45, and 145. Similarly, (b1)-(b4), (c1)-(c4), and (d1)-(d4) are the generated DMV-JND images, original images, and the DMV-JND distorted images in epoch 1, 15, 45, and 145.}\label{Subjective}
	\end{center}
\end{figure*}

Fig. \ref{Subjective} shows the CAM maps, DMV-JND images, original images, and DMV-JND distorted images, from the left to the right. Each column contains four results corresponding to epoch 1, 15, 45, and 145. The white regions in CAM maps suggests the low redundancy regions for the DMV, while the dark regions suggest the high redundancy regions. In the DMV-JND images, gray color with (125, 125, 125) values in RGB color space suggests no DMV-JND added in, while the other colors indicate different levels of the DMV-JND. As the epoch number increases, the gray color regions in the DMV-JND images become less, and this means larger DMV-JND is generated with the decreasing of $Loss2$. 
Meanwhile, the spatial distribution of the gray regions in the DMV-JND images is adjusted to those indicated by white regions in CAM maps gradually. This demonstrates that $Loss3$ can well regulate the spatial distribution of the generated DMV-JND, although $Loss3$ increases due to more noise being generated. Therefore, the DMV-JND can be well-restrained during the DMV-JND generation. All the results above indicate that our loss function is designed reasonably in modeling the DMV-JND. 

\subsection{WGN Test}
\label{IV-E}
For comparison, we add the same amount of WGN to the original images and generate the WGN-distorted images. The distorted images are fed to the classifiers, we get a much lower training RCA (i.e., 15.5\%). In other words, adding random noise (WGN) leads to the lower RCA, while adding the same amount of the DMV-JND via the DMV-JND-NET can achieve 100\% training RCA as already shown in Fig. \ref{RCA}.  

\subsection{Homogeneous Property Test}
\label{IV-F}
As mentioned in Section \ref{Intro}, the HVS-JND has the homogeneous property: the HVS cannot perceive any changes under the HVS-JND. In this subsection, we verify that our proposed DMV-JND has the similar homogeneous property: as the added noise is reduced from the found DMV-JND to zero, the DMV cannot ``perceive'' any changes in terms of RCA.

To this end, we first get the well trained DMV-JND-NET by fixing all the parameters of the DMV-JND model when the training RCA reaches 100\%. Then, we generate 8 new below-DMV-JND images $e_1$, $e_2$, ..., $e_8$ for $e$, which has the 8/9, 7/9, ..., 1/9 times of pixel value in $e$; More specifically, assume that the value of the $k^{th}$ pixel in $e$ is represented with $v_k$. Then, the $k^{th}$ pixel value in $e_1$, $e_2$, ..., $e_8$ can be represented as $v_{1, k}=\frac{8}{9} v_{k}$, $v_{2, k}=\frac{7}{9} v_{k}$, ..., $v_{8, k}=\frac{1}{9} v_{k}$. This process can be regarded as the noise reduction from our learned DMV-JND. For the CIFAR-10 dataset, there are 50,000 training images, with corresponding 50,000 DMV-JND images. We feed 50,000 DMV-JND images, 8$\times$50,000 new below-JND images, and their associated 50,000 original images to the Classifiers and get the RCA being 100\%. Therefore, our generated DMV-JND has the similar homogeneous property to that of the HVS-JND. All the noise below the DMV-JND can be tolerated by the DMV.

% \textcolor{blue}{It should be noticed that the traditional HVS-JND is a result that determined by most subjects (75\%). In view of this, the performance of the DMV task can be reduced in some degree. E.g., the RCA of the DMV-JND distorted images can be lower than 100\% during our modeling for the DMV-JND, also the as reduced noise from the achieved DMV-JND to zero the RCA can be increased in some degree, which is similar with the satisfied user ratio (SUR, the fraction of users that do not perceive any distortion when comparing the original image to its distorted version) \cite{zhang2020satisfied}.}  

\section{Conclusion}
In this paper, we have demonstrated that the Deep Machine Vision (DMV) has the just noticeable difference (JND), termed as the DMV-JND. We first define the concept of the DMV-JND. Then, the problem of the DMV-JND is carefully formulated. After that, we build the first JND model for the DMV. It can be achieved by the proposed DMV-JND-NET via unsupervised learning. To better restrain the DMV-JND generation, a semantic-guided redundancy assessment strategy is proposed and integrated into the DMV-JND-NET. Experimental results demonstrate that we successfully find and model the JND for the DMV. Additionally, we also highlight the potential applications of the DMV-JND, which exemplifies the DMV-oriented image and video processing.

\end{document}